\documentclass[conference]{IEEEtran}
\IEEEoverridecommandlockouts
\usepackage{cite}
\usepackage{amsmath,amssymb,amsfonts}
\usepackage{algorithmic}
\usepackage{graphicx}
\usepackage{textcomp}
\usepackage{xcolor}
\def\BibTeX{{\rm B\kern-.05em{\sc i\kern-.025em b}\kern-.08em
    T\kern-.1667em\lower.7ex\hbox{E}\kern-.125emX}}
\begin{document}

\title{Leveraging Exogenous Signals for Hydrology Time Series Forecasting}

\author{\IEEEauthorblockN{Junyang He}
\IEEEauthorblockA{\textit{College of Computing} \\
\textit{Georgia Institute of Technology}\\
Atlanta, GA, USA \\
jhe382@gatech.edu}
\and
\IEEEauthorblockN{Judy Fox}
\IEEEauthorblockA{\textit{School of Data Science} \\
\textit{University of Virginia}\\
Charlottesville, VA, USA \\
ckw9mp@virginia.edu}
\and
\IEEEauthorblockN{Alireza Jafari}
\IEEEauthorblockA{\textit{Computer Science} \\
\textit{University of Virginia}\\
Charlottesville, VA, USA \\
jrp5td@virginia.edu}
\and
\IEEEauthorblockN{Ying-Jung Chen}
\IEEEauthorblockA{\textit{College of Computing} \\
\textit{Georgia Institute of Technology}\\
Atlanta, GA, USA \\
yingjungcd@gmail.com}
\and
\IEEEauthorblockN{Geoffrey Fox}
\IEEEauthorblockA{\textit{Biocomplexity Institute} \\
\textit{University of Virginia}\\
Charlottesville, VA, USA \\
vxj6mb@virginia.edu}
}

\maketitle

\begin{abstract}
  Recent advances in time series research facilitate the development of foundation models. While many state-of-the-art time series foundation models have been introduced, few studies examine their effectiveness in specific downstream applications in physical science. This work investigates the role of integrating domain knowledge into time series models for hydrological rainfall-runoff modeling. Using the CAMELS-US dataset, which includes rainfall and runoff data from 671 locations with six time series streams and 30 static features, we compare baseline and foundation models. Results demonstrate that models incorporating comprehensive known exogenous inputs outperform more limited approaches, including foundation models. Notably, incorporating natural annual periodic time series contribute the most significant improvements. 
\end{abstract}

\section{Introduction}
\subsection{Spatio-temporal series}
Scientific data is frequently represented as spatio-temporal series, where time series data are often influenced by geographical factors. In hydrology, daily precipitation and streamflow are grouped by catchments and affected by environmental attributes and locations of these sites. The spatio-temporal nature of scientific time series data reflects strong dependence on the spatial variability of environmental factors. Hundreds of studies in recent years have proposed new deep learning models for time series analysis, largely concentrating on temporal dependencies \cite{sciencefmhub, ddz, liang2024foundation, neuralforecast2022}. Most general-purpose time series models overlook the critical influence of application-specific environmental and exogenous factors. This study not only demonstrates how embedding domain knowledge improves forecasting in hydrological applications, but also proposes a broader blueprint for science-specific foundation models. By rethinking the limitations of the ‘one-model-for-all’ paradigm, we outline how future foundation models can integrate physical laws, domain attributes, and exogenous drivers to pave the way for a new class of adaptive and interpretable models tailored to scientific discovery.

\subsection{Rainfall-runoff problem}

Rainfall–runoff models that tightly couple physical process representations with machine learning consistently improve streamflow simulation across diverse hydroclimates \cite{Konapala2020, Kratzert2019WRR}. By embedding learnable components into HBV-like structures, differentiable process-based models generate interpretable states (e.g., soil moisture, evapotranspiration, baseflow) while rivaling the predictive power of leading neural networks \cite{Feng2022WRR}. Process-driven deep learning approaches like \textsc{PRNN–EA–LSTM} integrate a compact conceptual core (EXP-HYDRO) into an entity-aware LSTM, strengthening low-flow and baseflow predictions without compromising interpretability \cite{Li2024JoH}. These hybrids leverage robust time series foundations: regional LSTM/EA-LSTM models demonstrated state-of-the-art results on CAMELS \cite{Kratzert2018HESS,Kratzert2019HESS}, whereas Transformer-family architectures, such as the Temporal Fusion Transformer, provide competitive multi-horizon forecasting with interpretable attention \cite{Lim2021IJF}.

\section{Methods}
\subsection{CAMELS dataset preprocessing}
This study uses the CAMELS-US dataset \cite{camels-us-2017}, which provides spatio-temporal data at 671 U.S. catchments, including time series (e.g., temperature, streamflow) and static attributes (e.g., land cover, soil aridity, spatial extent). We selected a 7,031-day interval from October 2, 1989, to December 31, 2008, aligning with the USGS water year \cite{dingman2015physical}. Time series variables were complete, while missing static features were imputed using attribute-wise means. The categorical feature “month” was ordinally encoded on a normalized $0 - 1$ scale to preserve temporal ordering.

\begin{figure*}
\centering
\includegraphics[width=0.6\linewidth]{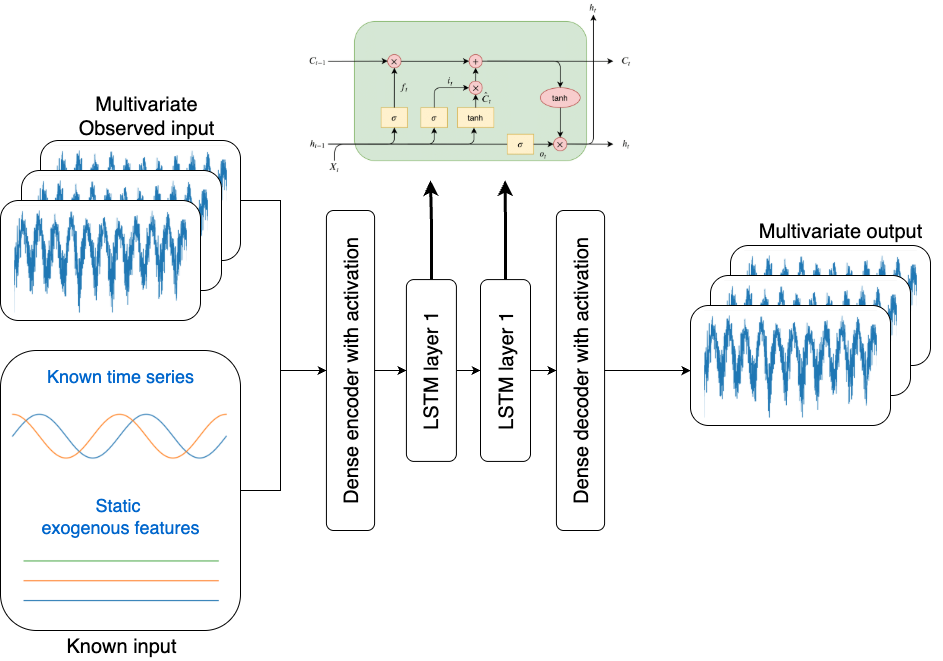}
\caption{LSTM model pipeline.}
\label{hydrology_lstm_architecture}
\end{figure*}

\subsection{Long Short-Term Memory}
As a baseline for rainfall–runoff prediction, we employ a Long Short-Term Memory (LSTM) network \cite{lstm-1997}, implemented in TensorFlow (Figure~\ref{hydrology_lstm_architecture}). SELU activations are applied to the dense encoder, decoder, and LSTM layers to promote stable gradient flow, while Sigmoid is used for recurrent activations to capture nonlinear gating dynamics. To reduce overfitting and improve generalization across catchments, a dropout rate of 20\% is applied throughout the network.

\subsection{Model training and evaluation}
Model inputs can be divided into known and observed features \cite{earthquake-2022}, analogous to exogenous regressors and autoregressive signals in machine learning. Known inputs are variables available for both past and future timesteps. In this study, they include static exogenous attributes (climatic and hydrologic signatures, and catchment topography) as well as seasonal patterns embedded in the time series. Observed inputs, by contrast, are only available historically and must be predicted forward; in hydrology, these include precipitation, temperature, and streamflow. The model outputs (targets) correspond to the forecasted hydrological variables over future horizons.

The input data are organized into batches with a sequence length of 21 days, selected after evaluating alternatives of 7, 14, and 365 days. The three-week horizon was found to balance efficiency with predictive skill, providing sufficient context to capture subtle hydrological dynamics while avoiding excessive sequence length.

\textbf{Spatial Validation.} We adopt a location-based validation strategy rather than temporal splitting, since catchments in the CAMELS-US dataset are spatially independent. The dataset is partitioned on an 8:2 ratio by location, with 534 of 671 catchments assigned to training and 137 to validation. This spatial split leverages the large number of available catchments while ensuring that evaluation reflects generalization to unseen basins.

\textbf{Training Details.} Each benchmark model is trained for up to 120 epochs, a value empirically chosen to balance predictive accuracy with computational cost. Training follows an early-improvement criterion, where an epoch is considered successful if either training or validation loss decreases relative to the previous epoch. Models are optimized with Adam \cite{kingma2017adammethodstochasticoptimization} (learning rate = 0.001) using mean-squared error as the loss function. For evaluation, we select the checkpoint at 120 epochs unless the loss curve indicates overfitting, in which case the best earlier checkpoint is used.

\textbf{Spatial and Temporal Encodings.} Similar to other scientific domains, hydrological time series exhibit strong seasonal regularities. For example, precipitation and streamflow at a given gauge location in October tend to follow recurring annual patterns. In addition, hydrological research has characterized approximate water residence times across reservoirs such as rivers, lakes, soils, and the atmosphere \cite{pidwirny_hydrologic_cycle}. To encode these known periodicities and dependencies, we incorporate a set of spatial and temporal encodings into the model during training:
\begin{enumerate}
\item \textbf{Linear space–time:} linear functions spanning the spatial extent of catchments and the temporal length of the series.
\item \textbf{Annual Fourier time:} sine and cosine functions with a one-year period.
\item \textbf{Extra Fourier time:} sine and cosine functions with shorter periods (8, 16, 32, 64, and 128 days) to capture higher-frequency variability.
\item \textbf{Legendre time:} Legendre polynomial functions (degrees 2–4) scaled to the total series length, capturing long-range temporal structure.
\end{enumerate}

\begin{figure*}
\centering
\includegraphics[width=\linewidth]{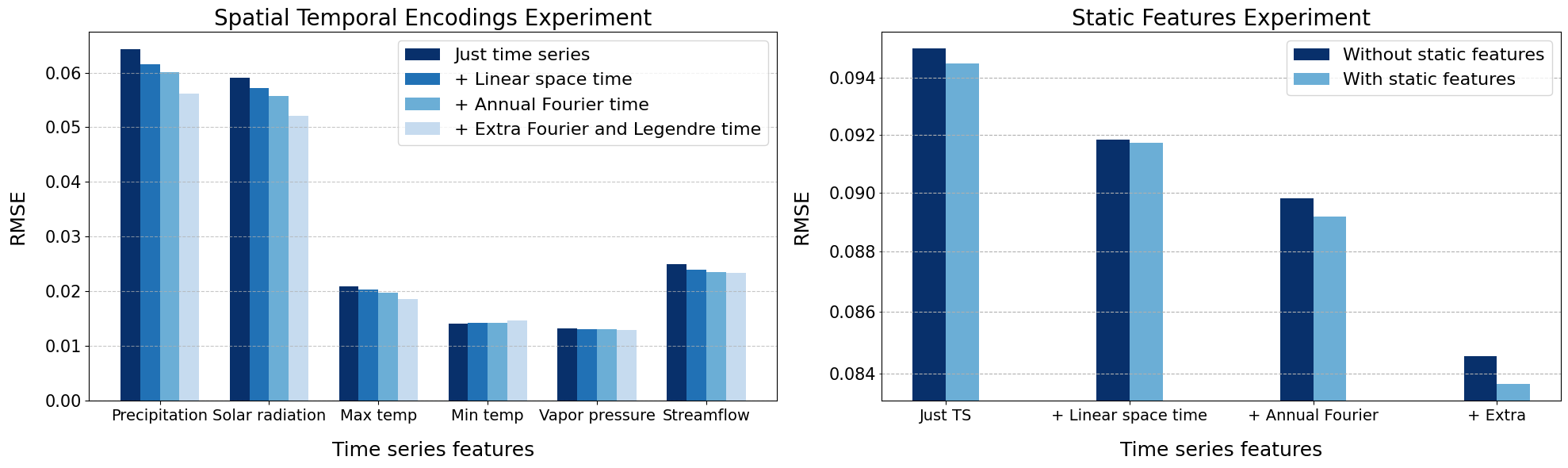}
\caption{Effects of spatial temporal encodings and static features (validation loss). \\
Left: RMSE by forecasted variable across four model configurations.
Right: RMSE by spatial–temporal encoding configuration (Just TS = time series only; Extra = additional Fourier + Legendre encodings)}
\label{static_encoding_experiment}
\end{figure*}

\begin{table*}[t]
\centering
\caption{Performance comparison of models}
\vspace{0.4em}
{\large               % <-- bigger font inside the group
\setlength{\tabcolsep}{6pt}    % default is ~6pt; increase for wider columns
\renewcommand{\arraystretch}{1.3} % >1.0 = taller rows
\begin{tabular}{l|c|c|c|c|c|c|c}
\hline
\textbf{Model} &
\textbf{In\&Out $^a$} &
\textbf{P $^b$} &
\textbf{SR $^c$} &
\textbf{Tmax $^d$} &
\textbf{Tmin $^e$} &
\textbf{VP $^f$} &
\textbf{Q $^g$} \\
\hline
LSTM-multivariate & MV\&MV & \textcolor{red}{0.0546} & \textcolor{red}{0.0509} &
\textcolor{red}{0.0138} & \textcolor{red}{0.0197} & \textcolor{red}{0.0122} & \textcolor{red}{0.0161} \\
Chronos-bolt-base & UV\&UV & 0.1103 & 0.0974 & 0.0332 & 0.0410 & 0.0466 & 0.0238 \\
Sundial-base-128m & UV\&UV & 0.1167 & 0.1033 & 0.0364 & 0.0446 & 0.0504 & 0.0252 \\
\hline
\multicolumn{8}{l}{\small $^a$UniVariate or MultiVariate training and forecasting $^b$Precipitation, $^c$Solar radiation, $^d$Maximum temperature.} \\
\multicolumn{8}{l}{\small $^e$Minimum temperature, $^f$Vapor pressure, $^g$Streamflow.} \\
\multicolumn{8}{l}{\small Table shows validation root mean squared error results. Lowest for each predicted target highlighted in red.} \\
\end{tabular}
} % end \large group
\label{foundation_model_experiment}
\end{table*}

\section{Benchmark experiments}
\subsection{Static features and encodings experiment}
We design a set of experiments to evaluate the contribution of known inputs—specifically seasonal climate patterns and static exogenous features—to forecasting performance. Consistent with the rainfall–runoff task, streamflow is excluded from training inputs and treated solely as a prediction target. Results show that spatial and temporal encodings consistently reduce root mean squared error (RMSE) across all six time series variables (Figure~\ref{static_encoding_experiment}, left). Moreover, incorporating static exogenous features yields additional improvements, further lowering RMSE (Figure~\ref{static_encoding_experiment}, right).

\subsection{Comparison with foundation models}
We benchmark our model against two general-purpose time series foundation models\footnote{https://github.com/JunyangHe/hydrology-foundation-models}: Chronos \cite{ansari2024chronoslearninglanguagetime} and Sundial \cite{liu2025sundialfamilyhighlycapable}. The models extend the benchmarking provided by He et al.\cite{doi:10.1177/10943420251380008}. Both are evaluated in a zero-shot setting, without task-specific fine-tuning. Performance results, summarized in Table~\ref{foundation_model_experiment}, are based on the same validation set comprising 137 catchments.

\textbf{LSTM-multivariate.} This model incorporates Linear Space, Linear Time, and Annual Fourier encodings, and is trained with all static and dynamic features, including streamflow. We adopt a multivariate training and forecasting setup \cite{doi:10.1080/02626669509491401}.

\textbf{Chronos-bolt-base.} Chronos \cite{ansari2024chronoslearninglanguagetime}, developed by Amazon, is a transformer-based model trained on tokenized time series. We use the bolt variant, the most recent and efficient release, which achieves the lowest reported error among Chronos models.

\textbf{Sundial-base-128m.} Sundial \cite{liu2025sundialfamilyhighlycapable}, introduced by Tsinghua University, is tailored for climate time series; more than 62\% of its pretraining corpus consists of ERA5 climate data \cite{era5-2021}.

Across all targets, our LSTM baseline outperforms both foundation models, demonstrating that simply scaling general-purpose models is insufficient; incorporating domain-specific knowledge remains essential for accurate and interpretable scientific time series forecasting.

\begin{figure*}
\centering
\includegraphics[width=\linewidth]{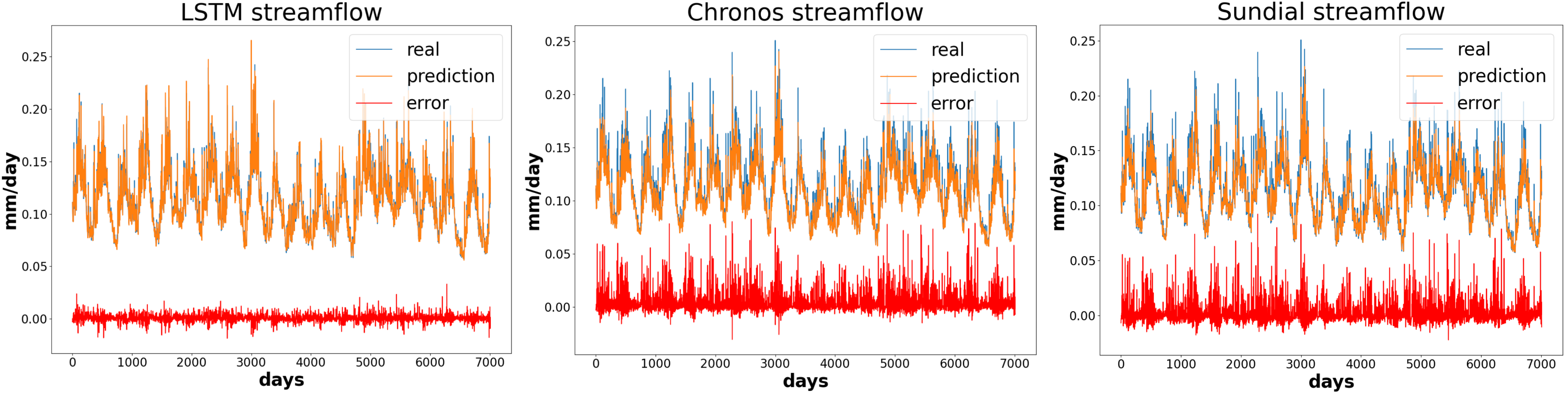}
\caption{Timeseriesviz package example output. Figure shows LSTM, Chronos-bolt, and Sundial model performances on streamflow data.}
\label{timeseriesviz}
\end{figure*}

\section{Time series plotting}

To support the broader research community working on spatio-temporal forecasting, we release timeseriesviz\footnote{https://github.com/JunyangHe/timeseriesviz}, an open-source Python package for scientific time series visualization. The library provides convenient plotting utilities for both NumPy arrays and Pandas DataFrames produced by NeuralForecast \cite{neuralforecast2022}, enabling consistent and reproducible analysis. An example visualization is shown in Figure~\ref{timeseriesviz}.

\section{Conclusion and Limitations}
This study shows that domain knowledge is critical for scientific time series forecasting. On the CAMELS rainfall–runoff task, an LSTM with multivariate training and known inputs outperforms state-of-the-art foundation models, challenging the “one-model-for-all” paradigm. These findings motivate domain-aware foundation models that couple general architectures with physical constraints and exogenous drivers, balancing accuracy with interpretability. Future work will examine how different forms of exogenous inputs shape forecast skill, moving toward adaptive and scientifically grounded foundation models.

\bibliographystyle{IEEEtran}
\bibliography{references}

\end{document}